\documentclass[sigconf]{acmart}

\AtBeginDocument{%
  }

\usepackage{wrapfig}
\usepackage{subcaption}

\usepackage{microtype}
\usepackage{graphicx}
\usepackage{booktabs} 
\usepackage{xcolor}
\usepackage{bm,bbm}
\usepackage{algorithm}
\usepackage{caption}
\usepackage[utf8]{inputenc} 
\usepackage[T1]{fontenc}    
\usepackage{hyperref}       
\usepackage{url}            
\usepackage{booktabs}       
\usepackage{amsfonts}       
\usepackage{nicefrac}       
\usepackage{microtype}      
\usepackage{xcolor}  
\usepackage{amsmath, amsthm, multirow, paralist}
\usepackage{mathtools}
\usepackage{bm,bbm}
\usepackage{hyperref}
\hypersetup{colorlinks=true,breaklinks=true}

\usepackage{algorithm}
\usepackage{listings}

\newtheorem*{theorem*}{Theorem}

\newcommand{\R}{{\mathbb{R}}}
\newcommand{\C}{{\mathbb{C}}}

\newcommand{\F}{{\mathcal{F}}}
\newcommand{\FF}{{\mathcal{F}^{-1}}}

\title{GCformer: An Efficient Framework for Accurate and Scalable Long-Term Multivariate Time Series Forecasting}

\author{YanJun Zhao}
\authornote{Three authors contributed equally to this research.}
\email{xiangyan.zyj@alibaba-inc.com}
\affiliation{%
  \institution{Xi'an Jiaotong University, Alibaba Group}
  \city{Xi'an}
  \country{China}
  \postcode{}
}

\author{Ziqing Ma}
\authornotemark[1]
\email{maziqing.mzq@alibaba-inc.com}
\affiliation{%
  \institution{Alibaba Group}
  \city{Hangzhou}
  \country{China}
  \postcode{}
}

\author{Tian Zhou}
\authornotemark[1]
\email{tian.zt@alibaba-inc.com}
\affiliation{%
  \institution{Alibaba Group}
  \city{Hangzhou}
  \country{China}
  \postcode{}
}

\author{Mengni Ye}
\email{mengni.yemengni@taobao.com}
\affiliation{%
  \institution{Alibaba Group}
  \city{Hangzhou}
  \country{China}
  \postcode{}
}

\author{Liang Sun}
\email{liang.sun@alibaba-inc.com}
\affiliation{%
  \institution{Alibaba Group}
  \city{Hangzhou}
  \country{China}
  \postcode{}
}

\author{Yi Qian}
\email{yqian@mail.xjtu.edu.cn}
\affiliation{%
\institution{Xi'an Jiaotong University}
\city{Xi'an}
\country{China}
\postcode{}
}

\begin{document}

  \begin{abstract}
 Transformer-based models have emerged as promising tools for time series forecasting. 
 However, these model cannot make accurate prediction for long input time series. On the one hand, they failed to capture long-range dependencies within time series data. On the other hand, the long input sequence usually leads to large model size and high time complexity. 
 To address these limitations, we present GCformer, which combines a structured global convolutional branch for processing long input sequences with a local Transformer-based branch for capturing short, recent signals. A cohesive framework for a global convolution kernel has been introduced, utilizing three distinct parameterization methods. The selected structured convolutional kernel in the global branch has been specifically crafted with sublinear complexity, thereby allowing for the efficient and effective processing of lengthy and noisy input signals. Empirical studies on six benchmark datasets demonstrate that GCformer outperforms state-of-the-art methods, reducing MSE error in multivariate time series benchmarks by 4.38\% and model parameters by 61.92\%. In particular, the global convolutional branch can serve as a plug-in block to enhance the performance of other models, with an average improvement of 31.93\%, including various recently published Transformer-based models. Our code is publicly available at https://github.com/zyj-111/GCformer.
  \end{abstract}




\maketitle


\section{Introduction}\label{sec_intro}

In light of the swift and successful development of Transformer within the fields of natural language processing (NLP) and computer vision (CV), an abundance of Transformer-based time series forecasting models have recently emerged~\cite{haoyietal-informer-2021,lim2020temporal}. However, these models suffer from some limitations, as they rely on a token-wise attention mechanism to construct a quadratic similarity matrix. Not only does this result in quadratic complexity in terms of memory and speed, but the vanilla Transformer also struggles with tasks involving long-term dependencies. Simply extending the input sequence does not lead to an increase of forecasting performance~\cite{LRA}. 

\begin{figure}[h]
\vskip -0.15in
\centering
\includegraphics[width=0.99\linewidth]{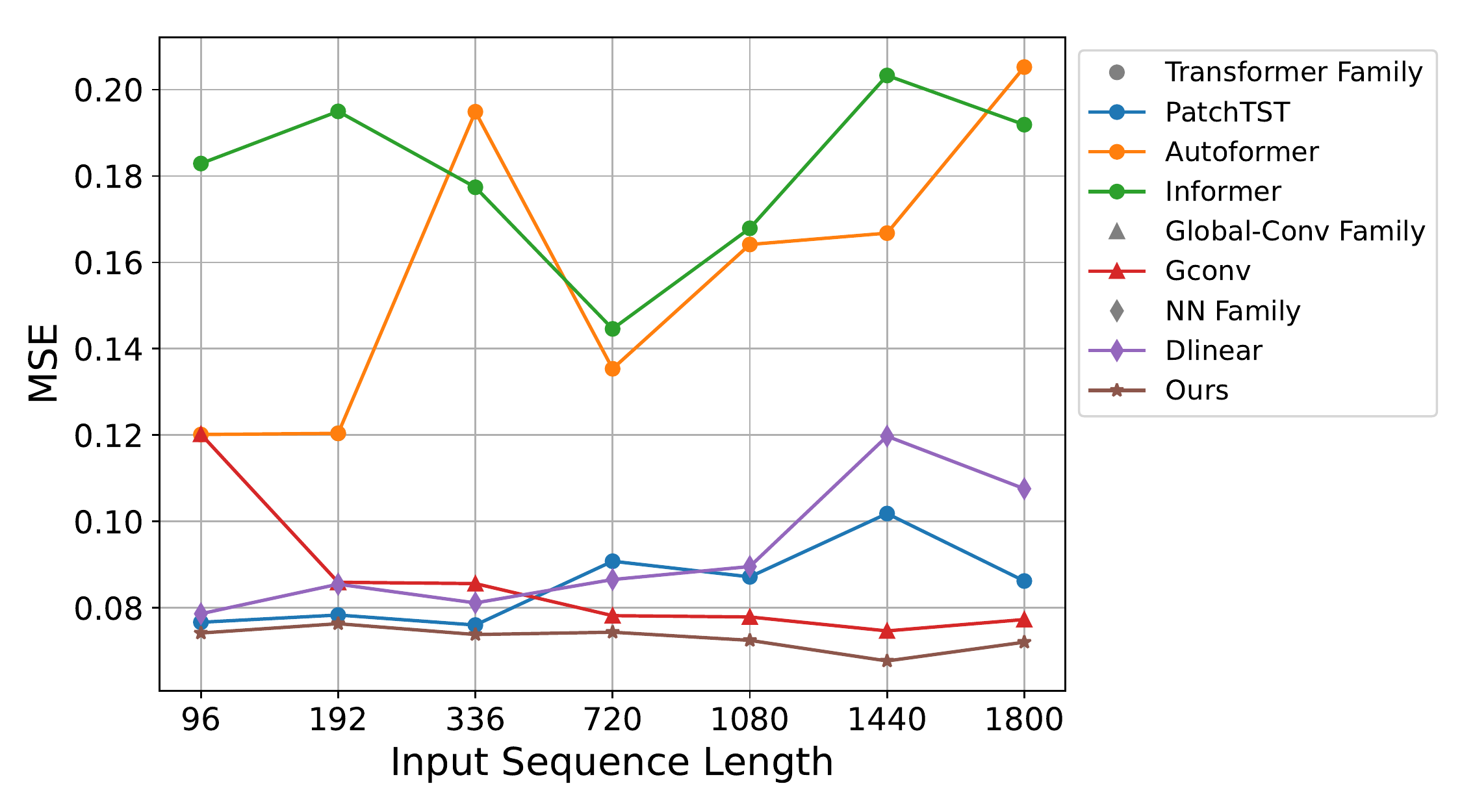}
\caption{Illustration of input length vs model performance (MSE: the lower the better) on a well-discussed time series benchmark. The experiment is performed on ETTm2 dataset in an univariate forecasting setting. The output length is fixed to be 96 and the input length varies in \{96, 192, 336, 720, 1080, 1440, 1800\}.}
\label{fig:length_compare}
\vskip -0.15in
\end{figure}

As illustrated in Figure~\ref{fig:length_compare}, we present how the forecasting performance (MSE) changes as the input length increases on a widely used forecasting benchmark dataset. 
Although it may seem intuitive that increasing the input length would result in richer historical information and improve performance, our studies suggest that this is not the case for many Transformer-based models, as well as some other neural network models. Similar results were reported in recent studies such as~\cite{haoyietal-informer-2021,Autoformer,FedFormer}. Researchers have developed long-dependency tasks: Long-range Arena (LRA)~\cite{LRA}, in addition to time series, to evaluate the performance of Transformers. These tasks involve sequences that can exceed 10k in length~\cite{LRA}. In response to these limitations, a new family of global convolutional networks has emerged, with the S4 model~\cite{S4} being a notable example. S4 is inspired by the state-space model and can be fast implemented as a global convolution kernel. Another study~\cite{Gconv} re-evaluated the global convolutional mechanism and emphasized the importance of employing a global kernel for long-sequence modeling. 

In contrast to canonical CNN models, which typically utilize a small reception field to capture local dependency, the global convolution approach employs a kernel that is equal in size to the length of the input sequence. This method is advantageous in terms of both speed and memory, as it allows for sublinear learnable parameterization with respect to input length and can be implemented using Fast Fourier Transform. The SGconv~\cite{Gconv} model has been shown to achieve SOTA performance on the LRA benchmark, and its success can be attributed to two key factors. First, the number of learnable parameters in the convolutional kernel scales sublinearly (logarithmically) to the input length. Second, the global kernel employs a weight decay property. 

Our research indicates that implementing a handcrafted weight decay rule may not be optimal for time series forecasting. This is because time series data often exhibits periodicity and a complex pattern that could benefit from a learnable kernel derived from the underlying data rather than simply relying on weight decay. This observation is consistent with that of Film~\cite{Film}, whose authors achieved SOTA performance using a no-weight-decaying Legendre measure. In this paper, we explore several methods for parameterizing a kernel with sublinear trainable parameters, including the use of multi-scale sub-kernels as proposed in~\cite{Gconv}, parameterization in the frequency domain, and parameterization in the high-order polynomial domain.

Here we summarize our key contributions as follows:

\begin{enumerate}
    \item We propose an enhanced global convolutional branch that can effectively capture long-term dependencies while sustaining sublinear parameter growth. 
    \item  Our proposed global convolutional branch represents a powerful tool that can be integrated as a plugin block to enhance diverse baseline algorithms. Our empirical findings show that it has the potential to substantially enhance the performance of local models by an average of {\bf 31.93\%}, while also achieving a remarkable {\bf 61.92\%} reduction in parameters. The optimal combination, GCformer, notably outperforms state-of-the-art methods by an impressive {\bf 4.38\%}.
    \item Our comprehensive framework introduces three parameterization methods that resemble global convolution techniques, revealing the versatility of our proposed paradigm when examining seemingly dissimilar algorithms.
\end{enumerate}

The remainder of this paper is structured as follows: In Section~\ref{sec_related_works}, we provide a summary of related work. Section~\ref{sec_model} introduces a generalized framework that employs diverse parameterization methods to analyze the global convolutional kernel. Additionally, we present the detailed model design for our proposed approach. In Section~\ref{sec_experiments}, we demonstrate the results of the numerical experiments in long-term time series forecasting benchmarks and conduct a comprehensive analysis to demonstrate the effectiveness of the plugin global convolution branch with various baseline local models. Furthermore, we present ablation experiments, parameter-saving analyses, speed comparisons, and robustness studies. 
Finally, in Section~\ref{sec_conclusion}, the conclusions and future research directions are discussed.


\section{Related Work}\label{sec_related_works}


\subsection{Transformer} 
The Transformer~\cite{attention_is_all_you_need} and its subsequent adaptations have demonstrated significant success in long sequence modelling tasks, including natural language processing (NLP)~\cite{attention_is_all_you_need,Bert/NAACL/Jacob}, time series anomaly detection~\cite{Anomaly-Transformer/iclr/XuWWL22,TFAD/cikm/ZhangZWS22}, and time series forecasting~\cite{haoyietal-informer-2021,DBLP:conf/iclr/KitaevKL20-reformer,liu2022pyraformer,Non-stationary-Transformers}, primarily due to its attention mechanism. This mechanism constructs a similarity matrix among time points, serving as an effective token mixer that captures dependencies within lengthy sequences. However, the Transformer model suffers from computation and memory complexity, which increases quadratically with input length, making it expensive to model long-range interactions in lengthy sequences. In addition, despite utilizing a long input coupled with abundant computational resources, many Transformer models still struggle to learn long-term dependencies and are prone to significant overfitting~\cite{LRA}. To overcome these challenges, several efficient variants of the Transformer model have been proposed recently~\cite{Efficient-Transformer-a-survey}. Some approaches sparsify the attention matrix to reduce parameters, such as using LogSparse attention~\cite{Log-transformer-shiyang-2019}, ProbSparse attention~\cite{haoyietal-informer-2021}, or using random pattern like Big Bird~\cite{zaheer2020bigbird}. Others approximate the attention matrix with fewer parameters, such as using low-rank approximation~\cite{DBLP:journals/corr/abs-2006-04768-linformer,xiong2021nystromformer}. Long-range Arena (LRA)~\cite{LRA} proposes a systematic and unified benchmark to evaluate performance under long-context scenarios, ranging from 1K to 16K tokens, which is significantly longer than that in time series tasks, like ETT benchmark~\cite{haoyietal-informer-2021}. 
Transformer-based models have proven less successful in that LRA benchmark.



\subsection{Convolutional models}
To efficiently capture long-term dependency, we might rethink using convolutional models. S4~\cite{S4} beats Transformer-based models on LRA benchmark. S4 is inspired by state-space model and can be viewed as a global convolution kernel. SGconv~\cite{Gconv} points out that a global convolutional kernel with sublinear parameterization and weight-decay structure may be the key to long-range dependency capturing. In fact, CNN has become popular in time series tasks~\cite{TCN,TCN2,TIMESNET} which uses one-dimensional convolution to extract local dependencies in time series. The effectiveness of CNNs is attributed to their use of convolution operations, which have inductive biases that introduce translation invariance. However, former CNNs focus on local details and the aggregation of information is enabled by stacking convolution layers, which is inefficient and fails to capture long-term dependencies~\cite{TIMESNET}.




\begin{figure*}[t]
\centering
\includegraphics[width=0.9\linewidth]{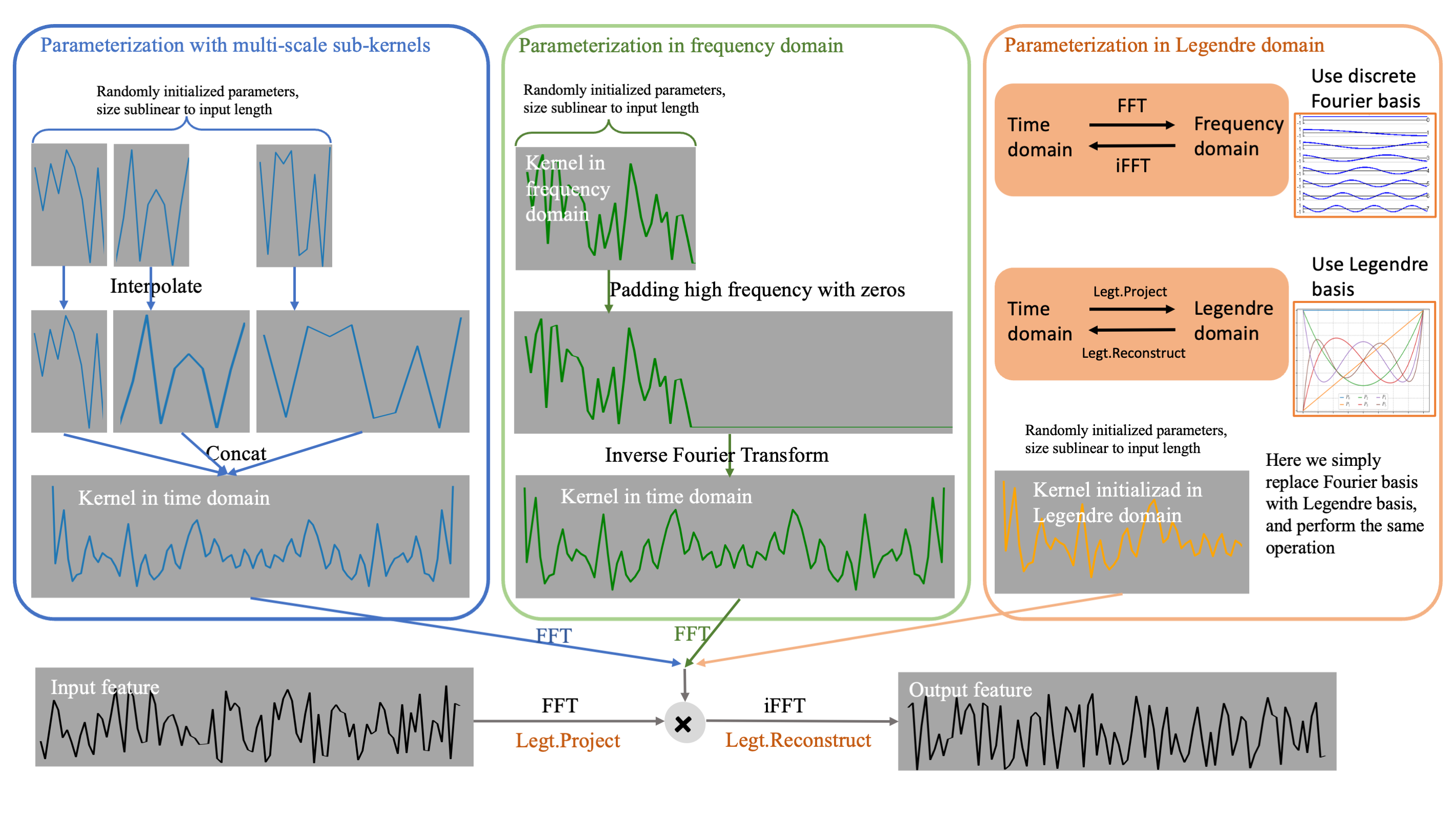}
\caption{Different parameterization methods of global convolution kernel.}
\label{fig:different_parameterization}
\end{figure*}

\subsection{Modeling both local and global context} 
The receptive field of a CNN is restricted by the size of the convolutional kernel, making it more suitable for local information modeling. On the other hand, Self-Attention (SA) computes attention weights for each position with respect to all positions, effectively taking into account the relationships between each point. Consequently, SA appears more appropriate for global modeling. However, point-wise representation might inhibit the model from summarizing a long historical signal, and long input SA may be prone to overfitting.

In order to capture both global and local information and model more granular information effectively, an increasing number of research works have adopted a hybrid approach by combining these two structures. MUSE~\cite{zhao2019muse} investigates the integration of convolution and self-attention mechanisms to enable the learning of sequence representations that incorporate multiple scales. T2VLAD~\cite{WangZ021} designs an efficient global-local alignment method to search relevant video contents based on natural language descriptions. VOLO~\cite{yuan2022volo} not only focuses on global dependency modeling at a coarse level, but also efficiently encodes finer-level features and contexts into token. In the field of time series forecasting, MICN~\cite{wang2023micn} proposes a local-global structure to implement information aggregation and long-term dependency modeling for time series, but both local information and global information are obtained through convolutional neural networks (CNNs), and two structures are composed in a serial manner.

\section{Method}
\label{sec_model}

In this section, we begin by providing a brief introduction of the global convolutional kernel, followed by the introduction of three distinct parameterization methods for the kernel.
We subsequently introduce a novel global-local architecture that leverages a low-complexity global branch to capture long-term dependencies and a Transformer-based local branch to efficiently capture fine-grained recent information. This innovative structure enhances the capacity of neural networks to model complex relationships in a computationally efficient manner.

\subsection{Global convolutional kernel}
The efficacy of a global convolution kernel in capturing long-term dependencies has been demonstrated in previous studies~\cite{Gconv}. Specifically, the authors employed a lengthy kernel that spanned the entire sequence length, thereby enables the capturing of long-term dependencies.
Given input sequence $u \in \R ^{n \times d}$, a learnable global kernel $k \in \R ^{n \times d}$, and output $y \in \R ^{n \times d}$. The global convolution kernel is represented as:
\begin{equation}
    y = u * k,
\end{equation}
where $*$ is the convolution operator. While the global convolution operation exhibits a complexity of $O(N^2)$, it can be fast implemented using Fast Fourier Transform denoted as $\F$, which has a complexity of $O(N \log N)$, leading to\begin{equation}
    u * k = \FF (\F(u) \cdot \F(k)). 
\end{equation}

\subsection{Efficient parameterization}
\paragraph{Parameterization with multi-scale sub-kernels}
A global convolutional kernel with parameters that scale linearly according to sequence length poses a significant challenge for efficient and effective feature extraction. To control the amount parameters only scales sublinearly to the sequence length, SGConv~\cite{Gconv} constructs the convolution kernel by assembling a series of sub-kernels with progressively larger sizes, where each sub-kernel is upsampled from an equal number of parameters using interpolation technology. Furthermore, they employed a weighted combination of sub-kernels where the weights are decaying, which can impart a favorable inductive bias for modeling extended sequences, leading to enhanced performance. Given a kernel $k_{\rm msk}$ parameterized with multi-scale sub-kernels, the process of global convolution model (${\rm Gconv_{msk}}$) is defined as
\begin{equation}
\label{func:gconv_msk}
    y = \FF (\F(u) \cdot \F (k_{\rm msk})). 
\end{equation}

\paragraph{Parameterization in frequency domain}
The distinct bias of time-frequency transform for time series data offers us an alternative way to parameterize a global convolution kernel. FEDformer~\cite{FedFormer} uses a compact representation of time series in frequency domain. Compared to SGConv, which generates the kernel in time domain, an alternative is to generate the kernel 
in frequency domain as $K_{\rm freq} \in \C ^{m \times d}$. To maintain a kernel with a sub-linear scale, we constrain $m << n$. The process of global convolution model (${\rm Gconv_{freq}}$) with a kernel parameterized in frequency domain is defined as
\begin{equation}
\label{func:gconv_freq}
    y = \FF (\F(u) \cdot K_{\rm freq}). 
\end{equation}

\begin{figure*}[t]
\centering
\includegraphics[width=0.9\linewidth]{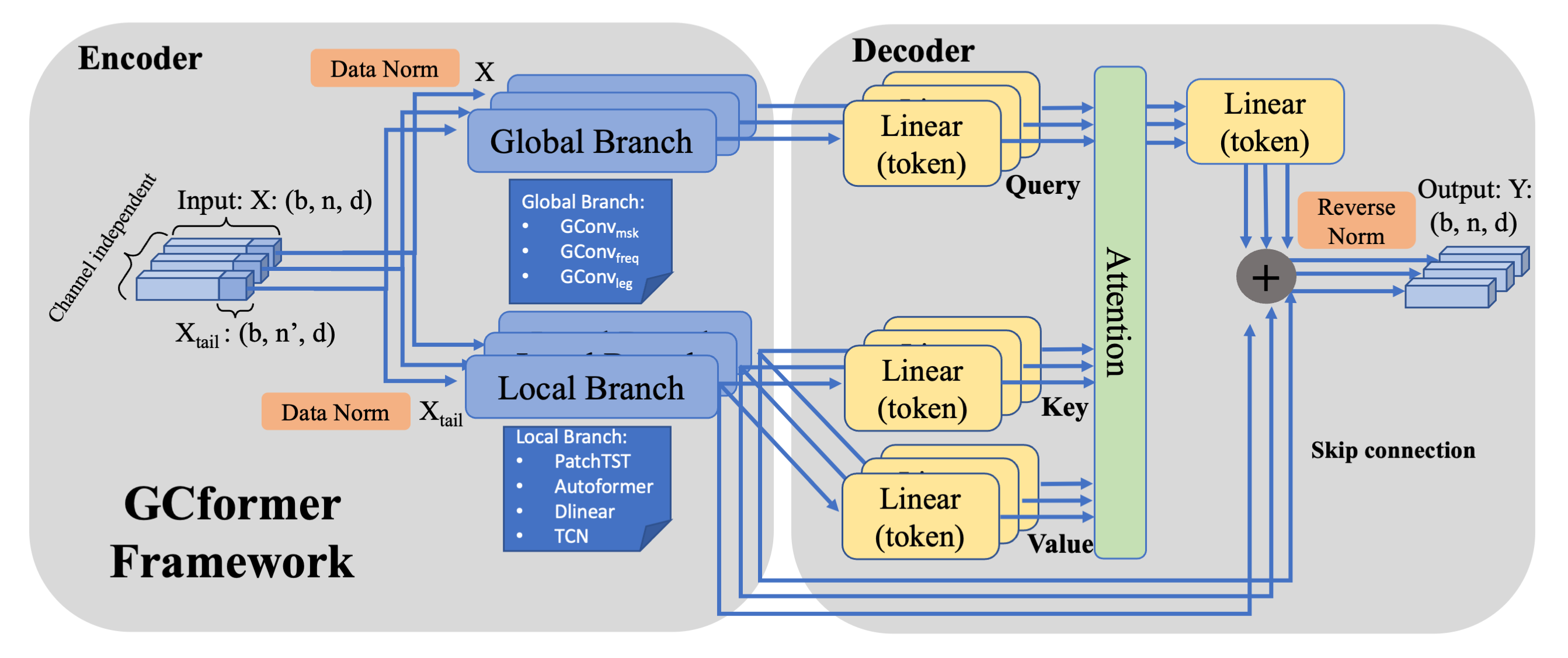}
\caption{GCformer overall framework.}
\label{fig:overall_structure}
\end{figure*}

\paragraph{From frequency domain to beyond} Recently, researchers have been utilizing state space models to process sequential data, achieving promising results. This approach offers us an alternative perspective for kernel parameterization beyond Fourier transform. The S4~\cite{S4} follows the process of state space model: 
$x_k = Ax_{k-1} + Bu_k, \ y_k = Cx_k + Du_k$, where $u_k \in \R ^{d}$ is the input signal at time step k, $x_k \in \R ^{d}$ is the hidden states of the state space model, and $y \in \R ^{d}$ is the output. The state transition matrix A and input matrix B is defined as:
\begin{align}
A_{n k}=(2 n+1) \begin{cases}(-1)^{n-k}, & \text { if } k \leq n \\ 1, & \text { if } k \geq n\end{cases}, 
B_{n}=(2 n+1)(-1)^{n}. 
\label{equ:LPU_A_B}
\end{align}
Here matrix $A \in \mathbb{R}^{d \times d }$ and $B \in \mathbb{R}^{d \times 1 }$ are derived with translated Legendre (LegT) measures which assign uniform weight to recent history~\cite{Hippo}. $C \in \mathbb{R}^{1 \times d }$ and $D \in \mathbb{R}^{1 \times 1 }$ are the output matrices.


The recurrent computation of the state-space model can be fast computed in a convolutional way. The transition matrices $A, B$, and $C$ are predefined so that the kernel $K$ can be computed in advance:
\begin{equation}
    y = u * K, \ K = (CB ,CAB,...,C A^{N-1} B).
\end{equation}
The transition matrix A is derived from Legendre Polynomials~\cite{Hippo}. Consequently, the operation $u * K$ can be interpreted as a projection of the original signal from time space to Legendre space. This process is similar to using Fourier Transform to project signal into frequency space. Using translated Legendre (LegT) measures, projecting a signal from time space to Legendre space is represented as: $\Bar{u} = {\rm LegT.Project}(u)$, and reconstructing a signal from Legendre space to time space is represented as: $u = {\rm LegT.Reconstruct}(\Bar{u})$. Given a kernel $K_{\rm leg} \in \R ^{m \times d}$, where $m << n$, the process of global convolution model (${\rm Gconv_{leg}}$) with translated Legendre measures is defined as:
\begin{equation}
\label{func:gconv_leg}
    y = {\rm legt.Reconstruct}({\rm legt.Project}(u) * K_{\rm leg}).
\end{equation}

\subsection{Synergistic Fusion of Global and Local branch}
\paragraph{Model structure} 


As shown in Figure \ref{fig:overall_structure}, our proposed architecture features a unique dual-branch design. This design involves simultaneously passing input data through two separate branches, each specifically designed to capture and extract distinct types of information: local and global. 
Our decoder module is specifically tailored to integrate and merge these two types of information in a manner that maximizes their complementarity. 

\paragraph{Encoder} Within the encoder part, there are two branches in parallel. Specifically, the upper branch is designed to extract global information $z_{\rm global}$, which refers to the long-term dependencies that exist within the sequence. Due to the expensive memory requirements of transformers, we address this issue by feeding the entire input sequence $X \in \mathbb{R}^{N \times d }$ into the global branch, whose complexity is sub-linear to the sequence length. In contrast, the lower branch focuses on capturing recent local information $z_{\rm local}$, which pertains to the dependencies between nearby time nodes. Meanwhile, we feed the tail segment of the sequence $X_{\rm tail} \in \mathbb{R}^{N' \times d }$ (where $N' < N$) to the Transformer branch to reduce the overall complexity without sacrificing prediction accuracy.
\begin{equation}
    z_{\rm global} = Branch_{\rm global}(X),
\end{equation}
\begin{equation}
     z_{\rm local} = Branch_{\rm local}(X_{tail}). 
\end{equation}
\paragraph{Decoder} 
To enhance the utilization of both global and local information, our approach involves incorporating the global information ($z_{\rm global}$) and local information ($z_{\rm local}$) into the decoder module, which outputs the prediction outcomes. The decoder module primarily comprises a cross-attention module aimed at ensuring the effective representation of historical information in time series. We map the global and local features to a hidden dimension at the token level, followed by utilizing the global information as the query ($\bm{q}$) and the local information as key ($\bm{k}$) and value ($\bm{v}$): 
\begin{equation}
    \bm{q} = {\rm MLP}(\bm{z}_{\rm global}),
\end{equation}

\begin{equation}
     \bm{k} = {\rm MLP}(\bm{z}_{\rm local}), \ \bm{v} = {\rm MLP}(\bm{z}_{\rm local}). 
\end{equation}
This allows for the effective integration of global and local information through querying the global information with the local information. 
\begin{equation}
\label{func:atten}
    {\rm Atten}(\bm{q},\bm{k},\bm{v})={\rm Softmax}(\frac{\bm{q}\bm{k}^\top}{\sqrt{d_q}})\bm{v}.
\end{equation}


\paragraph{Channel Independence} Channel-independence refers to the assumption in a dataset that there are no correlations between multiple variables. This concept was introduced in a previous study~\cite{patchTST} and has demonstrated improvements in several datasets. We also incorporate this idea into our work here.

\paragraph{Data Normalization} To address the issue of the distribution shift between the training and testing data, a data normalization method RevIN \cite{reversible} is utilized to enhance the robustness of the model. We first calculate the mean and standard deviation for each instance $x_{k}^{(i)} \in \mathbb{R}^{T}$ of input data:
\begin{equation}
    \mathbb{E}_{t}\left[x_{k t}^{(i)}\right]=\frac{1}{T} \sum_{j=1}^{T} x_{k j}^{(i)},
\end{equation}

\begin{equation}
    \operatorname{Var}\left[x_{k t}^{(i)}\right]=\frac{1}{T} \sum_{j=1}^{T}\left(x_{k j}^{(i)}-\mathbb{E}_{t}\left[x_{k t}^{(i)}\right]\right)^{2}.
\end{equation}

With these statistics, we normalize the input data using learnable affine parameter vectors $\gamma, \beta \in \mathbb{R}^{K}$ by
\begin{equation}
    \hat{x}_{k t}^{(i)}=\gamma_{k}\left(\frac{x_{k t}^{(i)}-\mathbb{E}_{t}\left[x_{k t}^{(i)}\right]}{\sqrt{\operatorname{Var}\left[x_{k t}^{(i)}\right]+\epsilon}}\right)+\beta_{k}. 
\end{equation}
The normalized data is then fed into the model for prediction. Finally, we reverse the normalization process by applying the reciprocal of the initial normalization to obtain the forecasting results.

\section{Experiments}
\label{sec_experiments}

\begin{table*}[h]
\centering
\vskip -0.15in
\caption{Results for boosting effect of global branch. We use PatchTST, Autoformer, Informer, Dlinear, and TCN as backbone models with input length=96. We leverage these backbone models with the global convectional branch (w G) and show the improvements compared to the original backbone (wo G) in the table. The experiments are performed in multivariate long-term series forecasting setting on seven benchmark datasets. A lower MSE indicates better performance. All experiments are repeated 3 times.}
\begin{center}
\begin{small}
\begin{sc}
\scalebox{0.75}{
\begin{tabular}{c|c|cccccccccccccccccc}
\toprule
\multicolumn{2}{c|}{Methods}&\multicolumn{2}{c|}{PatchTST(96)} & \multicolumn{2}{c|}{Dlinear(96)} &\multicolumn{2}{c|}{Autoformer}&\multicolumn{2}{c|}{Informer}&\multicolumn{2}{c|}{TCN}  & {Gconv}\\
\midrule
\multicolumn{2}{c|}{MSE w/wo G} & w G & wo G & w G & wo G & w G & wo G & w G & wo G & w G & wo G &  &\\
\midrule

\multirow{5}{*}{\rotatebox{90}{ETTm1}}
& 96  & 0.306 & 0.344  & \textbf{0.304} & 0.346 & 0.310 & 0.481 & 0.310 & 0.458 & 0.314 & 0.359  & 0.366 \\
& 192 & 0.348 & 0.370 & \textbf{0.337} & 0.383 & 0.344 & 0.628 & 0.352 & 0.564 & 0.351 & 0.388  & 0.355 \\
& 336 & 0.393 & 0.398 & \textbf{0.384} & 0.416 & 0.397 & 0.728 & 0.391 & 0.672 & 0.392 & 0.431  & 0.392 \\
& 720 & 0.436 & 0.461 & \textbf{0.429} & 0.478 & 0.466 & 0.658 & 0.453 & 0.714 & 0.434 & 0.499  & 0.452 \\
& improvement & \multicolumn{2}{c|}{$\uparrow 6.79\%$} & \multicolumn{2}{c|}{$\uparrow 10.52\%$} & \multicolumn{2}{c|}{$\uparrow 38.82\%$} & \multicolumn{2}{c|}{$\uparrow 37.06\%$} & \multicolumn{2}{c|}{$\uparrow 11.03\%$}   \\
\midrule

\multirow{5}{*}{\rotatebox{90}{ETTm2}}

& 96  & 0.173 & 0.179 & \textbf{0.172} & 0.186 & 0.176 & 0.255 & 0.176 & 0.365 & 0.175 & 3.041  & 0.192 \\
& 192 & 0.235 & 0.242 & \textbf{0.230} & 0.283 & 0.232 & 0.281 & 0.238 & 0.533 & 0.236 & 3.072  & 0.263 \\
& 336 & 0.296 & 0.306 & \textbf{0.284} & 0.376 & 0.315 & 0.339 & 0.310 & 1.363 & 0.294 & 3.105  & 0.363 \\
& 720 & 0.381 & 0.405 & \textbf{0.377} & 0.531 & 0.432 & 0.422 & 0.400 & 3.379 & 0.399 & 3.135  & 0.437 \\
& improvement & \multicolumn{2}{c|}{$\uparrow 3.85\%$} & \multicolumn{2}{c|}{$\uparrow 19.93\%$} & \multicolumn{2}{c|}{$\uparrow 13.26\%$} & \multicolumn{2}{c|}{$\uparrow 68.13\%$} & \multicolumn{2}{c|}{$\uparrow 91.09\%$}  \\

\midrule
\multirow{5}{*}{\rotatebox{90}{Electricity}} 
 
& 96  & \textbf{0.136} & 0.161 & 0.137 & 0.200 & 0.144 & 0.201 & 0.142 & 0.274 & 0.142 & 0.985  & 0.189 \\
& 192 & \textbf{0.152} & 0.172 & 0.159 & 0.199 & 0.171 & 0.222 & 0.168 & 0.296 & 0.174 & 0.996  & 0.199 \\
& 336 & \textbf{0.181} & 0.190 & 0.185 & 0.212 & 0.187 & 0.231 & 0.190 & 0.300 & 0.184 & 1.000  & 0.215 \\
& 720 & \textbf{0.234} & 0.241 & 0.250 & 0.249 & 0.246 & 0.254 & 0.271 & 0.373 & 0.247 & 1.438  & 0.260 \\
& improvement & \multicolumn{2}{c|}{$\uparrow 8.69\%$} & \multicolumn{2}{c|}{$\uparrow 15.98\%$}  & \multicolumn{2}{c|}{$\uparrow 18.38\%$} & \multicolumn{2}{c|}{$\uparrow 38.85\%$} & \multicolumn{2}{c|}{$\uparrow 83.13\%$}  \\

\midrule
\multirow{5}{*}{\rotatebox{90}{Traffic}} 
 
& 96  & \textbf{0.377} & 0.599 & 0.383 & 0.661 & 0.607 & 0.613 & 0.517 & 0.719 & 0.529 & 1.438  & 0.693 \\
& 192 & \textbf{0.393} & 0.455 & 0.395 & 0.612 & 0.503 & 0.616 & 0.461 & 0.696 & 0.455 & 1.463  & 0.673 \\
& 336 & \textbf{0.420} & 0.460 & 0.429 & 0.626 & 0.522 & 0.622 & 0.447 & 0.777 & 0.431 & 1.479  & 0.680 \\
& 720 & \textbf{0.445} & 0.490 & 0.459 & 0.645 & 0.474 & 0.660 & 0.467 & 0.864 & 0.484 & 1.499  & 0.773 \\
& improvement & \multicolumn{2}{c|}{$\uparrow 17.14\%$} & \multicolumn{2}{c|}{$\uparrow 34.46\%$}  & \multicolumn{2}{c|}{$\uparrow 15.89\%$} & \multicolumn{2}{c|}{$\uparrow 37.56\%$} & \multicolumn{2}{c|}{$\uparrow 67.46\%$} \\

 \midrule
\multirow{5}{*}{\rotatebox{90}{Weather}} 

& 96  & \textbf{0.156} & 0.176 & 0.176 & 0.198 & 0.180 & 0.266 & 0.183 & 0.300 & 0.178 & 0.615  & 0.179 \\
& 192 & \textbf{0.198} & 0.219 & 0.219 & 0.239 & 0.218 & 0.307 & 0.219 & 0.598 & 0.220 & 0.629  & 0.230 \\
& 336 & \textbf{0.254} & 0.274 & 0.271 & 0.286 & 0.278 & 0.359 & 0.265 & 0.578 & 0.267 & 0.639  & 0.288 \\
& 720 & \textbf{0.317} & 0.350 & 0.324 & 0.349 & 0.326 & 0578  & 0.320 & 1.059 & 0.320 & 0.639  & 0.384 \\
& improvement & \multicolumn{2}{c|}{$\uparrow 9.41\%$} & \multicolumn{2}{c|}{$\uparrow 7.97\%$} & \multicolumn{2}{c|}{$\uparrow 31.87\%$} & \multicolumn{2}{c|}{$\uparrow 56.58\%$} & \multicolumn{2}{c|}{$\uparrow 61.05\%$}   \\

 \midrule
\multirow{5}{*}{\rotatebox{90}{Illness}} 

& 24  & \textbf{1.533}  & 1.765 & 1.944 & 2.812 & 2.046 & 3.483 & 2.257 & 5.764 & 1.993 & 6.624  & 2.426 \\
& 36  & 1.625 & 1.713 & 1.598 & 2.691 & 2.052 & 3.103 & \textbf{1.543} & 4.755 & 1.938 & 6.858  & 2.542 \\
& 48  & \textbf{1.469}  & 1.589 & 2.089 & 2.810 & 2.159 & 2.669 & 2.021 & 4.763 & 2.094 & 6.968  & 2.486 \\
& 60  & \textbf{1.594} & 1.740 & 1.871 & 2.960 & 2.331 & 2.770 & 1.886 & 5.264 & 2.060 & 7.127  & 2.223 \\
& improvement & \multicolumn{2}{c|}{$\uparrow 8.55\%$} & \multicolumn{2}{c|}{$\uparrow 33.48\%$}  & \multicolumn{2}{c|}{$\uparrow 27.51\%$} & \multicolumn{2}{c|}{$\uparrow 62.52\%$} & \multicolumn{2}{c|}{$\uparrow 70.67\%$} \\

\midrule
\multirow{5}{*}{\rotatebox{90}{Exchange}}
 
& 96  & \textbf{0.0808}   & 0.0824 & 0.083 & 0.082 & 0.108 & 0.197 & 0.116 & 0.847 & 0.087 & 3.004  & 0.121 \\
& 192 & 0.169   & 0.167 & 0.199 & \textbf{0.161} & 0.236 & 0.300 & 0.248 & 1.204 & 0.202 & 4.048  & 0.302 \\
& 336 & 0.328    & 0.328 & 0.341 & \textbf{0.275} & 0.446 & 0.509 & 0.537 & 1.672 & 0.364 & 3.113  & 0.591 \\
& 720 & 0.879    & 0.956 & 1.192 & \textbf{0.710} & 1.18  & 1.447 & 1.484 & 2.478 & 1.237 & 3.150  & 1.801\\
& improvement & \multicolumn{2}{c|}{$\uparrow 2.43\%$} & \multicolumn{2}{c|}{$\downarrow -29.18\%$} & \multicolumn{2}{c|}{$\uparrow 24.33\%$} & \multicolumn{2}{c|}{$\uparrow 68.42\%$} & \multicolumn{2}{c|}{$\uparrow 85.28\%$}  \\
\midrule

\multicolumn{2}{c|}{Average improvement} & \multicolumn{2}{c|}{$\uparrow 8.12\%$} & \multicolumn{2}{c|}{$\uparrow 13.31\%$} & \multicolumn{2}{c|}{$\uparrow 24.29\%$}  & \multicolumn{2}{c|}{$\uparrow 52.73\%$}  & \multicolumn{2}{c|}{$\uparrow 76.45\%$}  
\\
\bottomrule
\end{tabular}

}
\label{tab:41GlobalBranchBoost}

\end{sc}
\end{small}
\end{center}
\vskip -0.2in
\end{table*}

\subsection{Dataset and implementation details}
\paragraph{Dataset} We briefly summarize the details of the datasets used in this article as follows: 1) ETT dataset~\cite{haoyietal-informer-2021} is collected from two separate counties in two versions of the sampling resolution (15 minutes \& 1 h). The ETT dataset contains several time series of electric loads and time series of oil temperature. 2) A dataset called Electricity\footnote{https://archive.ics.uci.edu/ml/datasets/ElectricityLoadDiagrams20112014} contains data on the electricity consumption of more than 300 customers and each column corresponds to the same client. 3) Traffic\footnote{http://pems.dot.ca.gov} dataset records the occupation rate of highway systems in California, USA. 4) The Weather\footnote{https://www.bgc-jena.mpg.de/wetter/} dataset contains 21 meteorological indicators in Germany for an entire year. 5) Illness\footnote{https://gis.cdc.gov/grasp/fluview/fluportaldashboard.html} dataset describes the number of patients with influenza-like diseases in the United States. 6) The Exchange~\cite{lai2018Modeling-exchange-dataset} dataset contains current exchanges of eight national currencies. Table~\ref{tab:dataset} summarizes all the features of the six benchmark datasets. They are divided into training sets, validation sets, and test sets in a 7:1:2 ratio during modeling.

\begin{table}[h]
\caption{Details of ETT benchmark datasets.}
\label{tab:dataset}
\begin{center}
\begin{small}
\begin{sc}
\begin{tabular}{l|cccr}
\toprule
Dataset & Length & Dimension & Frequency \\
\midrule
ETTm1/m2 & 69680$\approx$2 years & 7 & 15 min\\
Electricity & 26304$\approx$3 years & 321 & 1h & \\
Traffic & 17544$\approx$2 years & 862 & 1h & \\
Weather & 52696$\approx$1 years & 21 & 10 min & \\
Illness & 966$\approx$18 years & 7 & 7 days\\
Exchange & 7588$\approx$20 years & 8 & 1 day\\

\bottomrule
\end{tabular}
\end{sc}
\end{small}
\end{center}
\vskip -0.3in
\end{table}
\begin{table*}[t]
\centering
\begin{sc}
\caption{Multivariate long-term series forecasting results on six datasets with various input length and prediction length $\in \{96,192,336,720\}$ (For ILI dataset, we set prediction length $ \in \{24,36,48,60\}$). A lower MSE indicates better performance. All experiments are repeated 3 times.}\vspace{-1mm}
\scalebox{0.7}{
\begin{tabular}{c|c|cccccccccccccccccc}

\toprule
\multicolumn{2}{c|}{Methods}&\multicolumn{2}{c|}{GCformer}&\multicolumn{2}{c|}{PatchTST}&\multicolumn{2}{c|}{MICN}&\multicolumn{2}{c|}{FEDformer}&\multicolumn{2}{c|}{Autoformer}&\multicolumn{2}{c|}{S4}&\multicolumn{2}{c|}{Informer}&\multicolumn{2}{c|}{LogTrans}&\multicolumn{2}{c}{Dlinear}\\
\midrule
\multicolumn{2}{c|}{Metric} & MSE  & MAE & MSE & MAE & MSE  & MAE & MSE  & MAE& MSE  & MAE& MSE  & MAE& MSE  & MAE & MSE  & MAE & MSE & MAE\\
\midrule
\multirow{4}{*}{\rotatebox{90}{ETTm2}} 

& 96 & \textbf{0.163} & \textbf{0.251} & \underline{0.166} & \underline{0.256} &0.179 & 0.275 &0.203 &0.287 &0.255  &0.339 &0.705 &0.690 &0.365  &0.453  &0.768  &0.642  & 0.167  &0.260    \\
& 192 & \textbf{0.217} & \textbf{0.290} & \underline{0.223} & \underline{0.296} &0.262 &0.326 &0.269  &0.328  &0.281 &0.340 &0.924 &0.692 &0.533  &0.563  &0.989  &0.757 &0.224  &0.303 \\
& 336 & \textbf{0.268} & \textbf{0.322} & \underline{0.274} & \underline{0.329} &0.305 &0.353 & 0.325 &0.366 &0.339  &0.372 &1.364 &0.877 &1.363&0.887  &1.334  &0.872  & 0.281  &0.342   \\
& 720 & \textbf{0.351} & \textbf{0.379} & \underline{0.362} & \underline{0.385} &0.389 &0.407 &0.421 &0.415 &0.422  &0.419 &0.877 &1.074 &3.379  &1.338 & 3.048 &1.328  & 0.397  &0.421  \\
\midrule
\multirow{4}{*}{\rotatebox{90}{Electricity}} 

& 96 & \underline{0.132} & \underline{0.228} &\textbf{0.129}  &\textbf{0.222} &0.164 &0.269 &0.183 &0.297 &0.201  &0.317 &0.304 &0.405 &0.274  &0.368  &0.258  &0.357 &0.140 &0.237    \\
& 192 & \underline{0.152} & \underline{0.248} &\textbf{0.147}  & \textbf{0.240} &0.177 &0.285 & 0.195 &0.308 &0.222  &0.334 &0.313 &0.413 &0.296 &0.386  &0.266 &0.368  &0.153  &0.249    \\
& 336 & \underline{0.168} & \underline{0.266} & \textbf{0.163} & \textbf{0.259} &0.193 &0.304 &0.212 &0.313 &0.231 &0.338 &0.290 &0.381 &0.300  &0.394  &0.280 &0.380  &0.169  & 0.267   \\
& 720 & \underline{0.214} & \underline{0.307} & \textbf{0.197} & \textbf{0.290} &0.212 &0.321 & 0.231 &0.343 &0.254  &0.361 &0.262 &0.344 &0.373  &0.439 &0.283  &0.376 &0.203  &0.301    \\
\midrule

\multirow{4}{*}{\rotatebox{90}{Exchange}} 

&96  & \underline{0.080} & \textbf{0.196} & {0.097} & {0.220} &0.102 &0.235 &0.139 &0.276 &0.197  &0.323 &1.292 &0.849 &0.847  &0.752  &0.968  &0.812   & \textbf{0.078} & \underline{0.197}   \\
& 192 & \underline{0.167} & \textbf{0.290} & {0.197} & {0.315} &0.172 &0.316 &0.256 &0.369 &0.300  &0.369 &1.631 &0.968 &1.204   &0.895 &1.040  &0.851  & \textbf{0.159} & \underline{0.292}  \\
& 336 & 0.315 & \underline{0.407} & {0.351} & {0.429} &\textbf{0.272} &\underline{0.407} &0.426 &0.464 &0.509  &0.524 &2.225 &1.145 &1.672  &1.036  &1.659  &1.081  & \underline{0.274} & \textbf{0.391}   \\
& 720 & 0.768 & 0.667 & {1.02} & {0.737} &\underline{0.714} &\underline{0.658} &1.090 &0.800 &1.447  &0.941 &2.521 &1.245 &2.478  &1.310  &1.941  &1.127  & \textbf{0.558} & \textbf{0.574}  \\
\midrule

\multirow{4}{*}{\rotatebox{90}{Traffic}} 

&96  & \underline{0.377} & \underline{0.256} & \textbf{0.360}& \textbf{0.249} &0.519 &0.309 &0.562 &0.349 &0.613  &0.388 &0.824 &0.514 &0.719  &0.391  &0.684  &0.384  &0.410  &0.282    \\
& 192 & \underline{0.393} & \underline{0.268} & \textbf{0.379} & \textbf{0.256} &0.537 &0.315 &0.562 &0.346 &0.616&0.382 &1.106 &0.672 &0.696 &0.379  &0.685  &0.390   &0.423  &0.287    \\
& 336 & \underline{0.414} & \underline{0.291} & \textbf{0.392} & \textbf{0.264} &0.534 &0.317  &0.570 &0.323 &0.622  &0.337 &1.084 &0.627 &0.777  &0.420  &0.733  &0.408  &0.436 &0.296     \\
& 720 & \underline{0.445} & \underline{0.313} & \textbf{0.432} &\textbf{0.386} &0.577 &0.325 &0.596 &0.368 &0.660  &0.408 &1.536 &0.845 &0.864  &0.472  &0.717  &0.396  &0.466  &0.315     \\
\midrule

\multirow{4}{*}{\rotatebox{90}{Weather}} 

& 96 & \textbf{0.145} & \textbf{0.198} & \underline{0.149} & \underline{0.198} &0.161 &0.229   &0.217  &0.296  &0.266  &0.336 &0.406 &0.444 &0.300  &0.384  &0.458  &0.490  &0.176  &0.237    \\
& 192 & \textbf{0.187} & \textbf{0.237} &\underline{0.194} & \underline{0.241} &0.220 & 0.281 &0.276  &0.336  &0.307  &0.367 &0.525 &0.527 &0.598  &0.544  &0.658  &0.589  &0.220  &0.282    \\
& 336 & \textbf{0.244} & \textbf{0.281} &\underline{0.245} &\underline{0.282} &0.278 &0.331 & 0.339  &0.380  &0.359  &0.395 &0.531 &0.539 &0.578  &0.523  &0.797  &0.652  &0.265  &0.319    \\
& 720 & \textbf{0.311} & \textbf{0.331} &\underline{0.314} & \underline{0.334} &0.311 &0.356 &0.403  &0.428 &0.578 &0.578 &0.419  &0.428  &1.059  &0.741  &0.869  &0.675   &0.323  &0.362    \\
\midrule

\multirow{4}{*}{\rotatebox{90}{Illness}} 

& 24 & \textbf{1.258} & \textbf{0.773} & \underline{1.319} & \underline{0.754} &2.684 &1.112 &2.203  &0.963  &3.483 &1.287 &4.631 &1.484 &5.764  &1.677  &4.480  &1.444  &2.215  &1.081    \\
& 36 & \textbf{1.251} & \textbf{0.765} & \underline{1.579} & \underline{0.870} &2.667 &1.068 &2.272  &0.976  &3.103  &1.148 &4.123 &1.348 &4.755  &1.467  &4.799  &1.467 &1.963  &0.963    \\
& 48 & \textbf{1.240} & \textbf{0.767} & \underline{1.553} & \underline{0.815} &2.558 &1.052 &2.209  &0.981  &2.669  &1.085 &4.066 &1.36 &4.763  &1.469  &4.800  &1.468  &2.120  &1.024    \\
& 60 & \textbf{1.347} & \textbf{0.805} & \underline{1.470} & \underline{0.788} &2.747 &1.110 &2.545  &1.061  &2.770  &1.125 &4.278 &1.41 &5.264  &1.564  &5.278  &1.560  &2.368  &1.096    \\
\bottomrule
\end{tabular}
\label{tab:multi-benchmarks}
}
\vskip -0.1in
\end{sc}
\end{table*}

\paragraph{Implementation} We use ADAM \cite{kingma_adam:_2017} optimizer with a learning rate of $1e^{-4}$ to $1e^{-3}$. We save models with the lowest loss in validation sets for the final test. Measurements are made using mean square error (MSE) and mean absolute error (MAE). All experiments are repeated 3 times and the mean of the metrics is reported as the final result. We use a NVIDIA V100 32GB GPU and all deep learning networks (implemented in Pytorch~\cite{NEURIPS2019_9015_pytorch}) can fit in a single GPU with different experimental conditions.

\subsection{Global branch boosting results}

To evaluate the proposed framework, we conduct a vast experiment on six popular real-world benchmark datasets, including traffic, energy, economics, weather, and disease, for long-term prediction. We employ some state-of-the-art (SOTA) backbone models: PatchTST~\cite{patchTST}, Autoformer~\cite{Autoformer}, Informer~\cite{haoyietal-informer-2021}, Dlinear~\cite{Dlinear}, and TCN~\cite{TCN} to demonstrate the boosting effect of the global branch. For better comparison, we follow the experimental settings of previous work~\cite{Autoformer,FedFormer} where the input length of the local branch is fixed at 96 while the global branch use a larger window (336) as input length to include more information. The prediction lengths are fixed to be 96, 192, 336, and 720, respectively. As shown in Table \ref{tab:41GlobalBranchBoost}, introducing a global branch brings improvements in 34/35 cases and an average reduction of 31.93\% in MSE. Among the backbone models, PatchTST is the best baseline. When PatchTST is combined with the global branch, an 8.12\% reduction in MSE is achieved.

\begin{figure*}[t]
\centering
\includegraphics[width=0.7\linewidth]{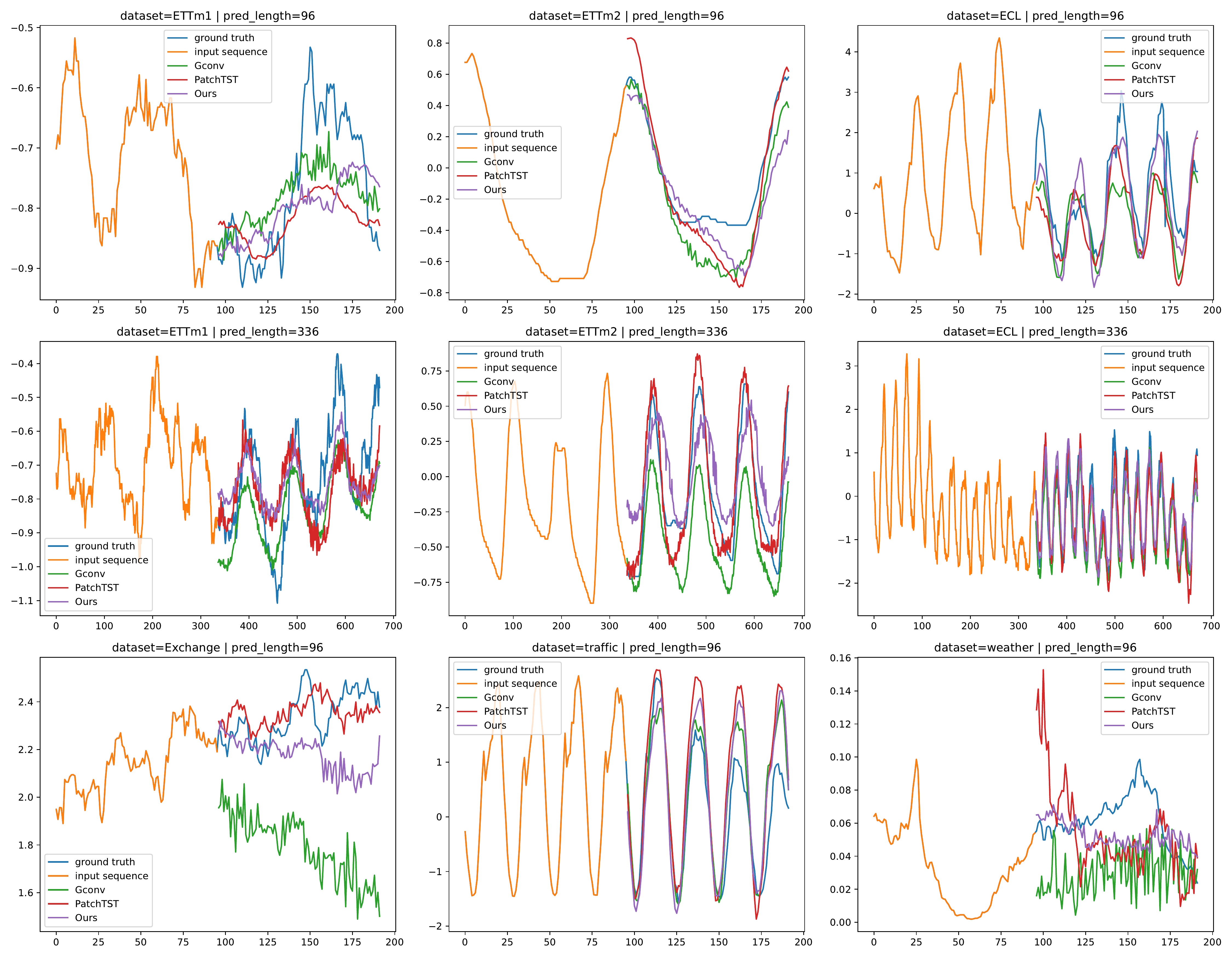}
\caption{Visualization of the forecasting outcomes of the global branch (Gconv), the local branch (PatchTST), and their combination (Ours). Experiments are conducted for 96-step and 336-step forecasting on ETTm1, ETTm2, Electricity, Exchange, Traffic, and Weather datasets. It can be observed that the combination of global and local branches effectively leverages the strengths of both and outperforms the use of a single branch.}
\label{fig:boost}
\vskip -0.15in
\end{figure*}

\subsection{Full benchmark}
For fair comparison, we follow the experimental settings of previous work, where the input length is tuned as a hyperparameter to achieve the best result. And the prediction lengths for both training and evaluation are fixed at 96, 192, 336, and 720, respectively.

In multivariate forecasting tasks,
GCformer achieves the best performance in the six benchmark datasets, as shown in Table~\ref{tab:multi-benchmarks}. The best combination of GCformer is Global(${\rm Gconv_{msk}}$)-Local(PatchTST). We add a global local model: MICN~\cite{micn} and a global convolutional model: S4~\cite{S4} as baseline models. Compared to SOTA work (PatchTST), our proposed GCformer yields an overall \textbf{4.38\%} relative MSE reduction. It is worth noting that the improvement is even more significant in some of the datasets, such as Illness (13.5\%). Our observations indicate that the GCformer outperforms the PatchTST model on datasets such as Illness and Exchange, which lack evident periodicity in their data. This is achieved through the GCformer's global convolutional branch, which effectively leverages long-term information. Alternatively, on datasets like Electricity and Traffic that exhibit clear periodicity, the PatchTST model demonstrates superior performance, and leveraging a global convolutional branch brings very limited improvements.



\subsection{Ablation study}
\subsubsection{Different global convolutional parameterization kernel}
To compare the impacts of various global convolution parameterization techniques, we conduct experiments on the global branch using a global convolutional kernel that is parameterized using three different methods. All three methods are combined with the same local Transformer branch for consistency in the comparison: ${\rm Gconv_{msk}}$ (Function~\ref{func:gconv_msk}), ${\rm Gconv_{freq}}$ (Function~\ref{func:gconv_freq}), and ${\rm Gconv_{leg}}$ (Function~\ref{func:gconv_leg}). As shown in Table~\ref{tab:different-gconv}, ${\rm Gconv_{msk}}$ outperforms the other two methods in most cases when combined with local features. This can be attributed to its more flexible learned parameterization pattern, as opposed to the other two methods that use fixed basis Fourier and Legendre functions.


\begin{table}[t]
\centering

\begin{sc}

\caption{Comparison of different global convolutional kernel (${\rm Gconv_{msk}}$, ${\rm Gconv_{freq}}$, ${\rm Gconv_{leg}}$) for multivariate long-term series forecasting on three representative datasets. A lower MSE indicates better performance. All experiments are repeated 3 times.}
\vskip -0.1in
\scalebox{0.8}{
\begin{tabular}{c|c|cccccccccccccccccc}
\toprule
\multicolumn{2}{c|}{Global branch}&\multicolumn{2}{c|}{${\rm Gconv_{msk}}$}&\multicolumn{2}{c|}{${\rm Gconv_{freq}}$}&\multicolumn{2}{c|}{${\rm Gconv_{leg}}$}\\
\midrule
\multicolumn{2}{c|}{Metric} & MSE  & MAE & MSE & MAE& MSE  & MAE\\
\midrule
\multirow{4}{*}{\rotatebox{90}{ETTm2}} 
&96  & \textbf{0.173} & \textbf{0.260} & 0.176 & 0.260 & 0.176 &0.259 \\
&192 & 0.235 & 0.300 & 0.236 & 0.303 & \textbf{0.231} & \textbf{0.298}  \\
&336 & 0.296 & 0.343 & \textbf{0.293} & \textbf{0.342} & 0.294 & 0.343   \\
&720 & \textbf{0.381} & \textbf{0.399} & 0.398 & 0.405 & 0.394 & 0.402   \\

\midrule
\multirow{4}{*}{\rotatebox{90}{Electricity}} 
&96  & \textbf{0.136} & \textbf{0.233} & 0.164 & 0.253 & 0.164 & 0.253  \\
&192 & \textbf{0.152} & \textbf{0.248} & 0.174 & 0.267 & 0.172 & 0.264  \\
&336 & \textbf{0.181} & \textbf{0.281} & 0.195 & 0.287 & 0.187 & 0.280 \\
&720 & 0.234 & 0.319 & 0.244 & 0.328 & \textbf{0.197} & \textbf{0.291}  \\

\midrule
\multirow{4}{*}{\rotatebox{90}{Weather}}
&96  & \textbf{0.156} & \textbf{0.205} & 0.173 & 0.214 & 0.172 & 0.214\\
&192 & \textbf{0.198} & \textbf{0.244} & 0.208 & 0.250 & 0.211 & 0.252  \\
&336 & \textbf{0.254} & \textbf{0.287} & 0.255 & 0.287 & 0.257 & 0.289  \\
&720 & \textbf{0.317} & \textbf{0.335} & 0.318 & 0.338 & 0.321 & 0.337 \\


\bottomrule
\end{tabular}
\label{tab:different-gconv}
}
\end{sc}
\vskip -0.1in
\end{table}

\subsubsection{Different structures in the decoder}
To leverage global and local information to the fullest extent, we have investigated a wide range of structures for the combination of the output of the local branch and the global branch.

1. Series Structure: We experiment with a series architecture that sequentially applies global and local branches step-by-step. Series-GL is defined as: 
$\bm{y}_{\rm pred} =  Branch_{\rm local}(Branch_{\rm global}(input))$. In this method, the input is initially fed into the global block to extract global information and then into the local block to extract local information. The reverse order (Series-LG) is also tested, which is defined as: $\bm{y}_{\rm pred} =  Branch_{\rm global}(Branch_{\rm local}(input))$.

2. Concatenate Structure: 
We also explore a straightforward approach, where the local and global information are concatenated in the sequence dimension: $\bm{z} = {\rm Concat}(\bm{z}_{global},\bm{z}_{local})$, where $\bm{z} \in \R^{2N \times d}$. Once combined, we apply an MLP to map $\bm{z}$ from length $2N$ to $N$: $\bm{y}_{\rm pred} = {\rm MLP}(\bm{z})$. With this approach, we utilize the MLP layer to determine the quality of predictions in both the local and global branches, assigning greater weight to those that are deemed better.

\begin{table}[h]
\centering
\caption{Comparison of different decoder structure for multivariate forecasting results on three representative datasets. A lower MSE indicates better performance. All experiments are repeated 3 times.}\vspace{-1mm}

\begin{center}
\begin{sc}
\scalebox{0.80}{
\begin{tabular}{c|c|cccccccccccccccc}
\toprule
\multicolumn{2}{c|}{Methods}&\multicolumn{2}{c|}{PatchTST(96)}&\multicolumn{2}{c|}{Attention}&\multicolumn{2}{c|}{Series-LG/GL}&\multicolumn{2}{c}{Concatenate}\\
\midrule
\multicolumn{2}{c|}{Metric} & MSE  & MAE & MSE & MAE& MSE  & MAE& MSE  & MAE\\
\midrule
\multirow{4}{*}{\rotatebox{90}{ETTm2}} 
&96 &0.179 & 0.261 & \textbf{0.173} & \textbf{0.260} & 0.179 & 0.265 & 0.178 & 0.261 \\
&192 &0.242 & 0.303 & \textbf{0.235} & \textbf{0.300} & 0.249 & 0.311 & 0.236 & 0.301 \\
&336 &0.306 & 0.342 & 0.296 & 0.343  & 0.315 & 0.352 & \textbf{0.286} & \textbf{0.338} \\
&720 &0.405 & 0.402 & 0.381 & 0.399 & 0.413 & 0.411 & \textbf{0.366} & \textbf{0.390} \\

\midrule
\multirow{4}{*}{\rotatebox{90}{Electricity}} 
&96 &0.161 & 0.249 & \textbf{0.136} & \textbf{0.233} & 0.173 & 0.268 & 0.166 & 0.256 \\
&192 &0.172 & 0.260 & \textbf{0.152} & \textbf{0.248} & 0.181 & 0.277 & 0.165 & 0.263 \\
&336 &0.190 & 0.279 & 0.181 & 0.281 & 0.203 & 0.298 & \textbf{0.177} & \textbf{0.276} \\
&720 &0.241 & 0.321 & 0.234 & 0.319 & 0.248 & 0.329 & \textbf{0.214} & \textbf{0.305} \\

\midrule
\multirow{4}{*}{\rotatebox{90}{Weather}}
&96 &0.176 & 0.219 & \textbf{0.156} & \textbf{0.205} & 0.179 & 0.221 & 0.175 & 0.217 \\
&192 &0.219 & 0.256 & \textbf{0.198} & \textbf{0.244} & 0.222 & 0.260 & 0.206 & 0.248 \\
&336 &0.274 & 0.295 & 0.254 & 0.287 & 0.277 & 0.298 & \textbf{0.251} & \textbf{0.286} \\
&720 &0.350 & 0.344 & 0.317 & 0.335 & 0.354 & 0.349 & \textbf{0.313} & \textbf{0.332} \\

\bottomrule
\end{tabular}
\label{tab:different-decoder}
}

\end{sc}
\end{center}
\vskip -0.1in
\end{table}

To compare the aforementioned structures and our proposed Attention structure in Function \ref{func:atten}, we conduct experiments on three representative datasets. As indicated in Table \ref{tab:different-decoder}, our findings demonstrate that the Attention and Concatenate architectures outperform other structures. Specifically, the Attention structure results in improvements of 3.85\%, 8.69\%, and 9.41\% on ETT, Electricity, and Weather respectively, while the Concatenate structure results in improvements of 4.80\%, 4.75\%, and 6.37\%. In all, the Attention structure outperforms the Concatenate structure with a 6.03\% improvement in all.

\subsubsection{Difference between modeling across token and across channel}
Multivariate time series data consists of several channels of variables. For instance, "ETTm2" contains 7 channels of variables, "Weather" contains 21 while "Electricity" comprises 321. To accurately predict future values, it is crucial to capture the relationship between the channels. 

We denote the shape of the time series data as (N,C), where N denotes the length of the token, and C signifies the number of channels in the data. As seen in most Transformers-based models, the time series data are usually handled in the token (N) dimension. However, in the Attention module in decoder of our proposed framework, we have devised two contrastive methods to handle it in the channel (C) dimension and the token (N) dimension, respectively. 

\begin{table}[h]
\begin{sc}

\caption{Comparison of different structure of Attention module in decoder. We compare the approaches of modelling over channel, token, and over both in decoder. A lower MSE indicates better performance on three typical datasets. All experiments are repeated 3 times on 3 representative datasets.}\vspace{-1mm}
\label{tab:token_channel}
\scalebox{0.8}{
\begin{tabular}{c|c|cccccccccccccccccc}
\toprule
\multicolumn{2}{c|}{Methods}&\multicolumn{2}{c|}{Token}&\multicolumn{2}{c|}{Channel}&\multicolumn{2}{c}{Both}\\
\midrule
\multicolumn{2}{c|}{Metric} & MSE  & MAE & MSE & MAE& MSE  & MAE \\
\midrule
\multirow{4}{*}{\rotatebox{90}{ETTm2}}
&96 &0.173 & 0.260 & \textbf{0.170} & \textbf{0.258} & 0.176 & 0.264  \\
&192 &0.235 & 0.300 & \textbf{0.231} & \textbf{0.303} & 0.239 & 0.307 \\
&336 &0.296 & 0.343 & \textbf{0.289} & \textbf{0.338} & 0.298 & 0.345 \\
&720 &0.381 & 0.399 & \textbf{0.369} & \textbf{0.390} & 0.394 & 0.406 \\
\midrule
\multirow{4}{*}{\rotatebox{90}{Electricity}}
&96  &\textbf{0.136} & \textbf{0.233} & 0.136 & 0.233 & 0.154 & 0.188  \\
& 192 &\textbf{0.152} & \textbf{0.248} & 0.154 & 0.250 & 0.188 & 0.279 \\
& 336 &\textbf{0.181} & \textbf{0.281} & 0.182 & 0.279 & 0.198 & 0.295 \\
& 720 &0.254 & 0.344 & \textbf{0.220} & \textbf{0.310} & 0.258 & 0.338 \\

\midrule
\multirow{4}{*}{\rotatebox{90}{Weather}}
&96  &\textbf{0.156} & \textbf{0.205} & 0.157 & 0.207 & 0.164 & 0.211  \\
& 192 &\textbf{0.198} & \textbf{0.244} & 0.198 & 0.244 & 0.200 & 0.247 \\
& 336 &0.254 & 0.287 & \textbf{0.252} & \textbf{0.286} & 0.254 & 0.287 \\
& 720 &0.317 & 0.335 & 0.315 & 0.334 & \textbf{0.314} & \textbf{0.334} \\
\bottomrule
\end{tabular}

}

\end{sc}
\vskip -0.1in
\end{table}
Based on the findings presented in Table~\ref{tab:token_channel}, it can be concluded that modeling in the token dimension is similarly effective as modeling in the channel dimension to predict short time series. However, when predicting lengthy time series, modeling in the channel dimension generates more precise predictions. We attribute this to the characteristics of the corresponding datasets that we plan to analyze in detail in future work. Nevertheless, simultaneous modeling in both the token and channel dimensions does not improve predictive accuracy due to overfitting. Although modeling in the channel dimension may offer better performance in some cases, it results in significant computational complexity. For example, modeling in the channel dimension requires more parameters than modeling in the token dimension when processing the Traffic dataset which has 862 channels. As a result, it may be more beneficial to model in the token dimension instead, especially when dealing with datasets that have a large number of channels.

\subsection{Model analysis}
\subsubsection{Learnable parameter size comparison} 
\begin{table}[h]
\begin{center}
\begin{sc}

\caption{Model parameter comparison for GCformer(336/192), GCformer(336/96), and PatchTST(336). We show the results on four representative datasets. The amount of parameters is expressed in millions (M).}
\vspace{-1mm}
\scalebox{0.8}{
\begin{tabular}{c|c|cccccccccccccccccc}
\toprule
\multicolumn{2}{c|}{Param (M)}  &{PatchTST} &{GCformer} &{reduction} &{GCformer} &{reduction}\\
\midrule

\multicolumn{2}{c|}{Input Len}  &336 & 336/192 & & 336/96 &\\

\midrule
\multirow{4}{*}{\rotatebox{90}{ETTm2}}
&96  &4.0    & 3.8    & $\downarrow4.48\%$  & 2.8   & $\downarrow30.20\%$ \\
&192  &7.6    & 6.0    & $\downarrow20.93\%$ & 3.9   & $\downarrow48.00\%$ \\
&336  &13.0   & 9.3    & $\downarrow28.53\%$ & 5.7   & $\downarrow56.22\%$ \\
&720  &27.5   & 18.1   & $\downarrow34.15\%$ & 10.3  & $\downarrow62.30\%$ \\

\midrule
\multirow{4}{*}{\rotatebox{90}{Electricity}}
&96   &166  & 106  & $\downarrow35.88\%$ & 59  & $\downarrow64.38\%$ \\
&192  &331  & 201  & $\downarrow39.32\%$ & 106 & $\downarrow67.85\%$ \\
&336  &580  & 343  & $\downarrow40.80\%$ & 177 & $\downarrow69.34\%$ \\
&720  &1243 & 722  & $\downarrow41.85\%$ & 367 & $\downarrow70.41\%$ \\

\midrule
\multirow{4}{*}{\rotatebox{90}{Weather}}
&96  &11.2   & 8.3    & $\downarrow25.76\%$ & 5.2    & $\downarrow53.31\%$ \\
&192  &22.0   & 14.6   & $\downarrow33.55\%$ & 8.4   & $\downarrow61.60\%$ \\
&336  &38.3   & 24.1   & $\downarrow36.98\%$ & 13.3  & $\downarrow65.25\%$ \\
&720  &81.7   & 49.4   & $\downarrow39.46\%$ & 26.2  & $\downarrow67.88\%$ \\

\midrule
\multirow{4}{*}{\rotatebox{90}{Traffic}}
&96  &445  & 281  & $\downarrow 36.79\%$ & 154 & $\downarrow 65.33\%$ \\
&192  &890  & 535  & $\downarrow39.80\%$ & 281 & $\downarrow 68.36\%$ \\
&336  &1557 & 917  & $\downarrow41.10\%$ & 472 & $\downarrow 69.65\%$ \\
&720  &3337 & 1935 & $\downarrow42.01\%$ & 981 & $\downarrow 70.58\%$ \\

\bottomrule
\end{tabular}

}
\label{tab:parameters}
\end{sc}
\end{center}
\vskip -0.1in
\end{table}

While the PatchTST model currently outperforms other time series models, its excessive parameter count and long training time become a bottleneck. In contrast, the convolutional kernel utilized in our proposed global branch is efficiently parameterized, thereby reducing overall number of parameters and scaling sub-linearly with sequence length. Consequently, this results in a faster training process and a lower memory consumption.
We use a Global(${\rm Gconv_{msk}}$)-Local(PatchTST) model as an example. By shortening the input length fed into the Local branch, our GCformer can significantly reduce its parameter count. To ensure the integrity of our experiment, we report the parameters of three models: GCformer(336/192) with input length=192 for local branch and input length=336 for global branch, GCformer(336/96) with input length = 96 for the local branch, and PatchTST with an input length of 336. As shown in Table \ref{tab:parameters}, GCformer(336/192) enjoys a lightweight property with a 33. 84\% reduction in learnable parameters compared to PatchTST, and GCformer (336/96) gives a 61. 92\% reduction on average.

\subsubsection{Robustness analysis}

\begin{table}[H]
\vskip -0.2in
\centering
\caption{Robustness analysis of multivariate results conducted on three typical datasets. The degree of noise injected into the time series data is determined by $\eta$. A lower MSE indicates better performance. All experiments are repeated 3 times.}\vspace{-1mm}
\begin{center}
\begin{small}
\begin{sc}

\scalebox{0.8}{
\begin{tabular}{c|c|cccccccccccccccccc}
\toprule
\multicolumn{2}{c|}{GCformer}&\multicolumn{2}{c|}{Original}&\multicolumn{2}{c|}{$\eta=1\%$}&\multicolumn{2}{c|}{$\eta=5\%$}&\multicolumn{2}{c}{$\eta=10\%$}\\
\midrule
\multicolumn{2}{c|}{Metric} & MSE  & MAE & MSE & MAE& MSE  & MAE& MSE & MAE\\
\midrule
\multirow{4}{*}{\rotatebox{90}{ETTm2}} 
&96  & 0.173 & 0.260 & \textbf{0.169} & \textbf{0.257} & 0.174 & 0.261 & 0.175 & 0.262  \\
&192 & 0.235 & \textbf{0.300} & \textbf{0.230} & \textbf{0.300} & 0.238 & 0.304 & \textbf{0.230} & 0.301  \\
&336 & 0.296 & 0.343 & 0.293 & 0.341 & 0.296 & 0.342 & \textbf{0.291} & \textbf{0.339}  \\
&720 & 0.381 & 0.399 & \textbf{0.375} & 0.398 & 0.388 & 0.400 & 0.384 & \textbf{0.392}  \\

\midrule
\multirow{4}{*}{\rotatebox{90}{Electricity}} 
&96  & \textbf{0.136} & \textbf{0.233} & 0.138 & 0.237 & 0.137 & 0.235 & 0.140 & 0.237  \\
&192 & \textbf{0.152} & \textbf{0.248} & 0.155 & 0.252 & 0.156 & 0.252 & 0.157 & 0.253 \\
&336 & 0.181 & 0.281 & \textbf{0.180} & 0.280 & 0.181 & \textbf{0.279} & 0.183 & \textbf{0.279} \\
&720 & 0.234 & 0.319 & \textbf{0.214} & \textbf{0.307} & 0.244 & 0.329 & 0.222 & 0.314  \\

\midrule
\multirow{4}{*}{\rotatebox{90}{Weather}}
&96  & \textbf{0.156} & \textbf{0.205} & 0.160 & 0.209 & 0.163 & 0.212 & 0.165 & 0.215   \\
&192 & \textbf{0.198} & \textbf{0.244} & 0.199 & \textbf{0.244} & 0.201 & 0.248 & 0.203 & 0.249 \\
&336 & \textbf{0.254} & \textbf{0.287} & 0.256 & \textbf{0.287} & 0.259 & 0.290 & 0.259 & 0.290 \\
&720 & 0.317 & 0.335 & 0.315 & \textbf{0.332} & \textbf{0.314} & \textbf{0.332} & 0.326 & 0.337\\


\bottomrule
\end{tabular}
\label{tab:robustness}
}
\end{sc}
\end{small}
\end{center}
\vskip -0.1in
\end{table}
To assess the robustness of our model, we employ a standard technique of injecting noise into the data and subsequently training the modified dataset. Then we record the performance metrics (MSE and MAE), which are presented in Table \ref{tab:robustness}. The results indicate a marginal increase in both metrics as the proportion of injected noise increases. For instance, on Weather dataset, compared to the raw experiment, the MSE increases by 0.81\%, 1.76\%, and 3.28\% when proportion of the noisy data is 1\%, 5\%, and 10\% respectively. This finding suggests that our proposed model is robust in handling data with low to moderate levels of noise (up to 10\%) and possesses a substantial advantage in dealing with various anomalous data. We also find that incorporation of small amounts of noise can have a positive impact on the performance on some noisy datasets, like ETTm2.



\subsubsection{Training speed}
\begin{figure}[t]
\centering
\includegraphics[width=0.99\linewidth]{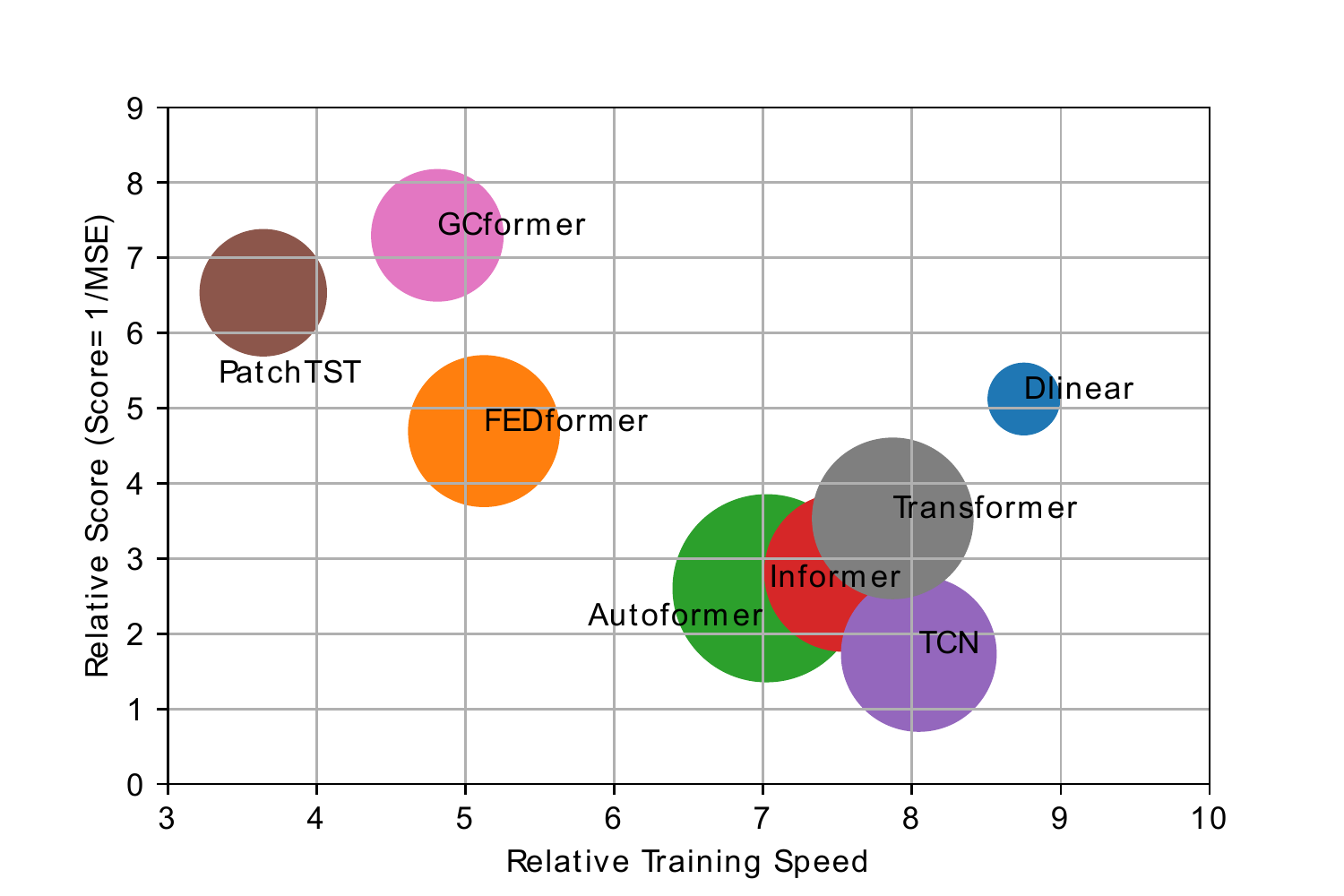}
\caption{Comparison of training speed and performance of benchmarks. The experiment is performed on "Electricity". The performance of the models is measured with Score, where Score = 1/MSE. The radius of the circle measures the standard deviation of MSE. A higher Score indicates better performance, same for Speed. A smaller circle indicates better robustness. The Speed and Score are presented on relative value.}
\label{fig:speed_score_scatter}
\vskip -0.2in
\end{figure}

Experiments are performed on one NVIDIA V100 32GB GPU. Figure \ref{fig:speed_score_scatter} displays the average speed versus the average performance in the benchmark setting. The results show that GCformer demonstrates significant advantages in terms of both speed and accuracy over its backbone local model PatchTST. Our model and PatchTST are slower than the other Transformer-based models due to the use of channel independent structure.

\section{Discussions and Conclusion}
\label{sec_conclusion}

In time series forecasting, the Transformer-based structures currently in use have limitations in capturing long-term dependencies and require large parameter size and training time due to their attention mechanism. We develop the global convolutional branch module to overcome this challenge, which effectively captures global information. We integrate it with a local attention-based module, resulting in a significant improvement in prediction accuracy.

Our extensive experiments demonstrate that our model can capture long-term dependencies more effectively and improve the model's overall performance on six benchmark datasets. Significantly, our proposed framework is quite versatile and can serve as a valuable built-in block for time series forecasting in future research. It can be easily adapted to various cases by replacing the local and global branches with other designed modules. Different parameterization methods of the global convolutional kernel are also worth exploring.




\bibliography{6_mybib}

\begin{thebibliography}{35}
\providecommand{\natexlab}[1]{#1}
\providecommand{\url}[1]{\texttt{#1}}
\expandafter\ifx\csname urlstyle\endcsname\relax
  \providecommand{\doi}[1]{doi: #1}\else
  \providecommand{\doi}{doi: \begingroup \urlstyle{rm}\Url}\fi

\bibitem[Bai et~al.(2018)Bai, Kolter, and Koltun]{TCN}
Bai, S., Kolter, J.~Z., and Koltun, V.
\newblock An empirical evaluation of generic convolutional and recurrent
  networks for sequence modeling.
\newblock \emph{CoRR}, abs/1803.01271, 2018.
\newblock URL \url{http://arxiv.org/abs/1803.01271}.

\bibitem[Devlin et~al.(2019)Devlin, Chang, Lee, and
  Toutanova]{Bert/NAACL/Jacob}
Devlin, J., Chang, M., Lee, K., and Toutanova, K.
\newblock {BERT:} pre-training of deep bidirectional transformers for language
  understanding.
\newblock In \emph{Proceedings of the 2019 Conference of the North American
  Chapter of the Association for Computational Linguistics: Human Language
  Technologies (NAACL-HLT), Minneapolis, MN, USA, June 2-7, 2019}, pp.\
  4171--4186, 2019.

\bibitem[Gu et~al.(2020)Gu, Dao, Ermon, Rudra, and R{\'e}]{Hippo}
Gu, A., Dao, T., Ermon, S., Rudra, A., and R{\'e}, C.
\newblock Hippo: Recurrent memory with optimal polynomial projections.
\newblock \emph{Advances in Neural Information Processing Systems},
  33:\penalty0 1474--1487, 2020.

\bibitem[Gu et~al.(2021)Gu, Goel, and R{\'e}]{S4}
Gu, A., Goel, K., and R{\'e}, C.
\newblock Efficiently modeling long sequences with structured state spaces.
\newblock \emph{arXiv preprint arXiv:2111.00396}, 2021.

\bibitem[Kim et~al.(2021)Kim, Kim, Tae, Park, Choi, and Choo]{reversible}
Kim, T., Kim, J., Tae, Y., Park, C., Choi, J.-H., and Choo, J.
\newblock Reversible instance normalization for accurate time-series
  forecasting against distribution shift.
\newblock In \emph{International Conference on Learning Representations}, 2021.

\bibitem[Kingma \& Ba(2017)Kingma and Ba]{kingma_adam:_2017}
Kingma, D.~P. and Ba, J.
\newblock Adam: {A} {Method} for {Stochastic} {Optimization}.
\newblock \emph{arXiv:1412.6980 [cs]}, January 2017.
\newblock arXiv: 1412.6980.

\bibitem[Kitaev et~al.(2020)Kitaev, Kaiser, and
  Levskaya]{DBLP:conf/iclr/KitaevKL20-reformer}
Kitaev, N., Kaiser, L., and Levskaya, A.
\newblock Reformer: The efficient transformer.
\newblock In \emph{8th International Conference on Learning Representations,
  {ICLR} 2020, Addis Ababa, Ethiopia, April 26-30, 2020}, 2020.

\bibitem[Lai et~al.(2018{\natexlab{a}})Lai, Chang, Yang, and Liu]{TCN2}
Lai, G., Chang, W., Yang, Y., and Liu, H.
\newblock Modeling long- and short-term temporal patterns with deep neural
  networks.
\newblock In Collins{-}Thompson, K., Mei, Q., Davison, B.~D., Liu, Y., and
  Yilmaz, E. (eds.), \emph{The 41st International {ACM} {SIGIR} Conference on
  Research {\&} Development in Information Retrieval, {SIGIR} 2018, Ann Arbor,
  MI, USA, July 08-12, 2018}, pp.\  95--104. {ACM}, 2018{\natexlab{a}}.
\newblock \doi{10.1145/3209978.3210006}.
\newblock URL \url{https://doi.org/10.1145/3209978.3210006}.

\bibitem[Lai et~al.(2018{\natexlab{b}})Lai, Chang, Yang, and
  Liu]{lai2018Modeling-exchange-dataset}
Lai, G., Chang, W.-C., Yang, Y., and Liu, H.
\newblock Modeling long-and short-term temporal patterns with deep neural
  networks.
\newblock In \emph{The 41st International ACM SIGIR Conference on Research \&
  Development in Information Retrieval}, pp.\  95--104, 2018{\natexlab{b}}.

\bibitem[Li et~al.(2019)Li, Jin, Xuan, Zhou, Chen, Wang, and
  Yan]{Log-transformer-shiyang-2019}
Li, S., Jin, X., Xuan, Y., Zhou, X., Chen, W., Wang, Y.-X., and Yan, X.
\newblock Enhancing the locality and breaking the memory bottleneck of
  transformer on time series forecasting.
\newblock In \emph{Advances in Neural Information Processing Systems},
  volume~32, 2019.

\bibitem[Li et~al.(2022)Li, Cai, Zhang, Chen, and Dey]{Gconv}
Li, Y., Cai, T., Zhang, Y., Chen, D., and Dey, D.
\newblock What makes convolutional models great on long sequence modeling?
\newblock \emph{CoRR}, abs/2210.09298, 2022.
\newblock \doi{10.48550/arXiv.2210.09298}.
\newblock URL \url{https://doi.org/10.48550/arXiv.2210.09298}.

\bibitem[Lim et~al.(2020)Lim, Arik, Loeff, and Pfister]{lim2020temporal}
Lim, B., Arik, S.~O., Loeff, N., and Pfister, T.
\newblock Temporal fusion transformers for interpretable multi-horizon time
  series forecasting, 2020.

\bibitem[Liu et~al.(2022{\natexlab{a}})Liu, Yu, Liao, Li, Lin, Liu, and
  Dustdar]{liu2022pyraformer}
Liu, S., Yu, H., Liao, C., Li, J., Lin, W., Liu, A.~X., and Dustdar, S.
\newblock Pyraformer: Low-complexity pyramidal attention for long-range time
  series modeling and forecasting.
\newblock In \emph{International Conference on Learning Representations},
  2022{\natexlab{a}}.

\bibitem[Liu et~al.(2022{\natexlab{b}})Liu, Wu, Wang, and
  Long]{Non-stationary-Transformers}
Liu, Y., Wu, H., Wang, J., and Long, M.
\newblock Non-stationary transformers: Rethinking the stationarity in time
  series forecasting.
\newblock \emph{CoRR}, abs/2205.14415, 2022{\natexlab{b}}.
\newblock \doi{10.48550/arXiv.2205.14415}.
\newblock URL \url{https://doi.org/10.48550/arXiv.2205.14415}.

\bibitem[Nie et~al.(2022)Nie, Nguyen, Sinthong, and Kalagnanam]{patchTST}
Nie, Y., Nguyen, N.~H., Sinthong, P., and Kalagnanam, J.
\newblock A time series is worth 64 words: Long-term forecasting with
  transformers.
\newblock \emph{CoRR}, abs/2211.14730, 2022.
\newblock \doi{10.48550/arXiv.2211.14730}.
\newblock URL \url{https://doi.org/10.48550/arXiv.2211.14730}.

\bibitem[Paszke et~al.(2019)Paszke, Gross, Massa, Lerer, Bradbury, Chanan,
  Killeen, Lin, Gimelshein, Antiga, Desmaison, Kopf, Yang, DeVito, Raison,
  Tejani, Chilamkurthy, Steiner, Fang, Bai, and
  Chintala]{NEURIPS2019_9015_pytorch}
Paszke, A., Gross, S., Massa, F., Lerer, A., Bradbury, J., Chanan, G., Killeen,
  T., Lin, Z., Gimelshein, N., Antiga, L., Desmaison, A., Kopf, A., Yang, E.,
  DeVito, Z., Raison, M., Tejani, A., Chilamkurthy, S., Steiner, B., Fang, L.,
  Bai, J., and Chintala, S.
\newblock Pytorch: An imperative style, high-performance deep learning library.
\newblock In \emph{Advances in Neural Information Processing Systems}, pp.\
  8024--8035. 2019.

\bibitem[Tay et~al.(2020)Tay, Dehghani, Abnar, Shen, Bahri, Pham, Rao, Yang,
  Ruder, and Metzler]{LRA}
Tay, Y., Dehghani, M., Abnar, S., Shen, Y., Bahri, D., Pham, P., Rao, J., Yang,
  L., Ruder, S., and Metzler, D.
\newblock Long range arena: A benchmark for efficient transformers.
\newblock \emph{arXiv preprint arXiv:2011.04006}, 2020.

\bibitem[Tay et~al.(2023)Tay, Dehghani, Bahri, and
  Metzler]{Efficient-Transformer-a-survey}
Tay, Y., Dehghani, M., Bahri, D., and Metzler, D.
\newblock Efficient transformers: {A} survey.
\newblock \emph{{ACM} Comput. Surv.}, 55\penalty0 (6):\penalty0 109:1--109:28,
  2023.
\newblock \doi{10.1145/3530811}.
\newblock URL \url{https://doi.org/10.1145/3530811}.

\bibitem[Vaswani et~al.(2017)Vaswani, Shazeer, Parmar, Uszkoreit, Jones, Gomez,
  Kaiser, and Polosukhin]{attention_is_all_you_need}
Vaswani, A., Shazeer, N., Parmar, N., Uszkoreit, J., Jones, L., Gomez, A.~N.,
  Kaiser, {\L}., and Polosukhin, I.
\newblock Attention is all you need.
\newblock \emph{Advances in neural information processing systems}, 30, 2017.

\bibitem[Wang et~al.(2023{\natexlab{a}})Wang, Peng, Huang, Wang, Chen, and
  Xiao]{micn}
Wang, H., Peng, J., Huang, F., Wang, J., Chen, J., and Xiao, Y.
\newblock {MICN}: Multi-scale local and global context modeling for long-term
  series forecasting.
\newblock In \emph{The Eleventh International Conference on Learning
  Representations}, 2023{\natexlab{a}}.
\newblock URL \url{https://openreview.net/forum?id=zt53IDUR1U}.

\bibitem[Wang et~al.(2023{\natexlab{b}})Wang, Peng, Huang, Wang, Chen, and
  Xiao]{wang2023micn}
Wang, H., Peng, J., Huang, F., Wang, J., Chen, J., and Xiao, Y.
\newblock {MICN}: Multi-scale local and global context modeling for long-term
  series forecasting.
\newblock In \emph{The Eleventh International Conference on Learning
  Representations}, 2023{\natexlab{b}}.
\newblock URL \url{https://openreview.net/forum?id=zt53IDUR1U}.

\bibitem[Wang et~al.(2020)Wang, Li, Khabsa, Fang, and
  Ma]{DBLP:journals/corr/abs-2006-04768-linformer}
Wang, S., Li, B.~Z., Khabsa, M., Fang, H., and Ma, H.
\newblock Linformer: Self-attention with linear complexity.
\newblock \emph{CoRR}, abs/2006.04768, 2020.

\bibitem[Wang et~al.(2021)Wang, Zhu, and Yang]{WangZ021}
Wang, X., Zhu, L., and Yang, Y.
\newblock {T2VLAD:} global-local sequence alignment for text-video retrieval.
\newblock In \emph{{IEEE} Conference on Computer Vision and Pattern
  Recognition, {CVPR} 2021, virtual, June 19-25, 2021}, pp.\  5079--5088.
  Computer Vision Foundation / {IEEE}, 2021.
\newblock \doi{10.1109/CVPR46437.2021.00504}.
\newblock URL
  \url{https://openaccess.thecvf.com/content/CVPR2021/html/Wang\_T2VLAD\_Global-Local\_Sequence\_Alignment\_for\_Text-Video\_Retrieval\_CVPR\_2021\_paper.html}.

\bibitem[Wu et~al.(2021)Wu, Xu, Wang, and Long]{Autoformer}
Wu, H., Xu, J., Wang, J., and Long, M.
\newblock Autoformer: Decomposition transformers with auto-correlation for
  long-term series forecasting.
\newblock In \emph{Proceedings of the Advances in Neural Information Processing
  Systems (NeurIPS)}, pp.\  101--112, 2021.

\bibitem[Wu et~al.(2022)Wu, Hu, Liu, Zhou, Wang, and Long]{TIMESNET}
Wu, H., Hu, T., Liu, Y., Zhou, H., Wang, J., and Long, M.
\newblock Timesnet: Temporal 2d-variation modeling for general time series
  analysis.
\newblock \emph{CoRR}, abs/2210.02186, 2022.
\newblock \doi{10.48550/arXiv.2210.02186}.
\newblock URL \url{https://doi.org/10.48550/arXiv.2210.02186}.

\bibitem[Xiong et~al.(2021)Xiong, Zeng, Chakraborty, Tan, Fung, Li, and
  Singh]{xiong2021nystromformer}
Xiong, Y., Zeng, Z., Chakraborty, R., Tan, M., Fung, G., Li, Y., and Singh, V.
\newblock Nystr{\"o}mformer: A nystr{\"o}m-based algorithm for approximating
  self-attention.
\newblock 2021.

\bibitem[Xu et~al.(2022)Xu, Wu, Wang, and
  Long]{Anomaly-Transformer/iclr/XuWWL22}
Xu, J., Wu, H., Wang, J., and Long, M.
\newblock Anomaly transformer: Time series anomaly detection with association
  discrepancy.
\newblock In \emph{The Tenth International Conference on Learning
  Representations, {ICLR} 2022, Virtual Event, April 25-29, 2022}.
  OpenReview.net, 2022.
\newblock URL \url{https://openreview.net/forum?id=LzQQ89U1qm\_}.

\bibitem[Yuan et~al.(2022)Yuan, Hou, Jiang, Feng, and Yan]{yuan2022volo}
Yuan, L., Hou, Q., Jiang, Z., Feng, J., and Yan, S.
\newblock Volo: Vision outlooker for visual recognition.
\newblock \emph{IEEE Transactions on Pattern Analysis and Machine
  Intelligence}, 2022.

\bibitem[Zaheer et~al.(2020)Zaheer, Guruganesh, Dubey, Ainslie, Alberti,
  Ontanon, Pham, Ravula, Wang, Yang, et~al.]{zaheer2020bigbird}
Zaheer, M., Guruganesh, G., Dubey, K.~A., Ainslie, J., Alberti, C., Ontanon,
  S., Pham, P., Ravula, A., Wang, Q., Yang, L., et~al.
\newblock Big bird: Transformers for longer sequences.
\newblock \emph{Advances in Neural Information Processing Systems}, 33, 2020.

\bibitem[Zeng et~al.(2022)Zeng, Chen, Zhang, and Xu]{Dlinear}
Zeng, A., Chen, M., Zhang, L., and Xu, Q.
\newblock Are transformers effective for time series forecasting?
\newblock \emph{CoRR}, abs/2205.13504, 2022.
\newblock \doi{10.48550/arXiv.2205.13504}.
\newblock URL \url{https://doi.org/10.48550/arXiv.2205.13504}.

\bibitem[Zhang et~al.(2022)Zhang, Zhou, Wen, and Sun]{TFAD/cikm/ZhangZWS22}
Zhang, C., Zhou, T., Wen, Q., and Sun, L.
\newblock {TFAD:} {A} decomposition time series anomaly detection architecture
  with time-frequency analysis.
\newblock In Hasan, M.~A. and Xiong, L. (eds.), \emph{Proceedings of the 31st
  {ACM} International Conference on Information {\&} Knowledge Management,
  Atlanta, GA, USA, October 17-21, 2022}, pp.\  2497--2507. {ACM}, 2022.
\newblock \doi{10.1145/3511808.3557470}.
\newblock URL \url{https://doi.org/10.1145/3511808.3557470}.

\bibitem[Zhao et~al.(2019)Zhao, Sun, Xu, Zhang, and Luo]{zhao2019muse}
Zhao, G., Sun, X., Xu, J., Zhang, Z., and Luo, L.
\newblock Muse: Parallel multi-scale attention for sequence to sequence
  learning, 2019.

\bibitem[Zhou et~al.(2021)Zhou, Zhang, Peng, Zhang, Li, Xiong, and
  Zhang]{haoyietal-informer-2021}
Zhou, H., Zhang, S., Peng, J., Zhang, S., Li, J., Xiong, H., and Zhang, W.
\newblock Informer: Beyond efficient transformer for long sequence time-series
  forecasting.
\newblock In \emph{The Thirty-Fifth {AAAI} Conference on Artificial
  Intelligence, {AAAI} 2021, Virtual Conference}, volume~35, pp.\
  11106--11115, 2021.

\bibitem[Zhou et~al.(2022{\natexlab{a}})Zhou, Ma, Wang, Wen, Sun, Yao, Yin, and
  Jin]{Film}
Zhou, T., Ma, Z., Wang, X., Wen, Q., Sun, L., Yao, T., Yin, W., and Jin, R.
\newblock Film: Frequency improved legendre memory model for long-term time
  series forecasting.
\newblock In \emph{NeurIPS}, 2022{\natexlab{a}}.
\newblock URL
  \url{http://papers.nips.cc/paper\_files/paper/2022/hash/524ef58c2bd075775861234266e5e020-Abstract-Conference.html}.

\bibitem[Zhou et~al.(2022{\natexlab{b}})Zhou, Ma, Wen, Wang, Sun, and
  Jin]{FedFormer}
Zhou, T., Ma, Z., Wen, Q., Wang, X., Sun, L., and Jin, R.
\newblock {FEDformer}: Frequency enhanced decomposed transformer for long-term
  series forecasting.
\newblock In \emph{39th International Conference on Machine Learning (ICML)},
  2022{\natexlab{b}}.

\end{thebibliography}
\bibliographystyle{neurips_2022}





\end{document}